\documentclass[sigconf]{acmart}
\usepackage{subfigure}
\usepackage{multirow}

\usepackage{algorithm}
\usepackage[noend]{algpseudocode}
\usepackage[normalem]{ulem}

\AtBeginDocument{%
  \providecommand\BibTeX{{%
    \normalfont B\kern-0.5em{\scshape i\kern-0.25em b}\kern-0.8em\TeX}}}

\copyrightyear{2024}
\acmYear{2024}
\setcopyright{rightsretained}
\acmConference[GECCO '24 Companion]{Genetic and Evolutionary Computation
Conference}{July 14--18, 2024}{Melbourne, VIC, Australia}
\acmBooktitle{Genetic and Evolutionary Computation Conference (GECCO '24
Companion), July 14--18, 2024, Melbourne, VIC, Australia}
\acmDOI{10.1145/3638530.3654265}
\acmISBN{979-8-4007-0495-6/24/07}

\begin{document}

\title{Lexidate: Model Evaluation and Selection with Lexicase}



 \author{Jose Guadalupe Hernandez}
 \orcid{0000-0002-1298-5551}
 \affiliation{%
   \institution{Cedars-Sinai Medical Center}
   \streetaddress{}
   \city{Los Angeles}
   \state{CA}
   \country{USA}
   \postcode{}
 }
 \email{jose.hernandez8@cshs.org}

 \author{Anil Kumar Saini}
 \orcid{0000-0002-9211-1079}
 \affiliation{%
   \institution{Cedars-Sinai Medical Center}
   \streetaddress{}
   \city{Los Angeles}
   \state{CA}
   \country{USA}
   \postcode{}
 }
 \email{anil.saini@cshs.org}

 \author{Jason H. Moore}
 \orcid{0000-0002-5015-1099}
 \affiliation{%
   \institution{Cedars-Sinai Medical Center}
   \streetaddress{}
   \city{Los Angeles}
   \state{CA}
   \country{USA}
   \postcode{}
 }
 \email{jason.moore@csmc.edu}

\renewcommand{\shortauthors}{Hernandez, Saini, and Moore}

\begin{abstract}

Automated machine learning streamlines the task of finding effective machine learning pipelines by automating model training, evaluation, and selection.
Traditional evaluation strategies, like cross-validation (CV), generate one value that averages the accuracy of a pipeline's predictions.
This single value, however, may not fully describe the generalizability of the pipeline.
Here, we present Lexicase-based Validation (lexidate), a method that uses multiple, independent prediction values for selection.
Lexidate splits training data into a learning set and a selection set.
Pipelines are trained on the learning set and make predictions on the selection set. 
The predictions are graded for correctness and used by lexicase selection to identify parent pipelines. 
Compared to $10$-fold CV, lexicase reduces the training time.
We test the effectiveness of three lexidate configurations within the Tree-based Pipeline Optimization Tool 2 (TPOT2) package on six OpenML classification tasks.
In one configuration, we detected no difference in the accuracy of the final model returned from TPOT2 on most tasks compared to 10-fold CV.
All configurations studied here returned similar or less complex final pipelines compared to 10-fold CV.

\end{abstract}



\maketitle

\section{Introduction}




Automated machine learning (AutoML) reduces the human intervention needed to build machine learning pipelines by automating the processes of building, training, evaluating, and selecting a pipeline \cite{waring2020automl}.
Multiple AutoML packages exist and each implements a unique approach to searching the space of possible pipelines. 
Some packages only require data to start, but computational limitations (\textit{e.g.}, runtime and hardware) must still be considered. 
Once a set of candidate pipelines is evaluated, the most promising ones are investigated further.
Numerous pipelines are examined with these tools, however, multiple replicates are needed to discover a diverse set of potential pipelines.
Cross-validation (CV) \cite{stone1974cv,geisser1975predictive,mosteller1968data} is the standard approach to maximize the information that can be extracted from small data sets \cite{hastie2009elements,bishop2006pattern}. 
With CV, training data is split into multiple partitions, and sequentially, one partition is used to assess a model's performance while the others are used to train the model.
This process continues until each partition has been used to assess a model, and the average is taken across the resulting scores.
While CV is an effective approach, numerous issues must still be considered.
For example, evaluating a large number of models with fixed partitions increases the risk of overfitting, the effectiveness of CV is problem-dependent, and the run time escalates with an increasing number of partitions  \cite{rao2008dangers,schaffer1993selecting,hastie2009elements,bishop2006pattern}.


Here, we introduce the lexicase-based validation (lexidate) strategy for model evaluation and selection within evolutionary AutoML systems.
Lexidate splits data into a learning set and a selection set.
Machine learning pipelines are trained on the learning set and make predictions on the selection set; predictions are graded for correctness.
The set of graded predictions is used to identify parents with lexicase selection \cite{helmuth2015lexicase}.
By using independent predictions to select parents, selection pressure can be applied to harder individual cases without sacrificing overall performance.
We compare the results of three lexidate configurations to $10$-fold CV when used within the Tree-Based Pipeline Optimization Tool 2 \cite{ribeiro2024tpot2} on six OpenML classification tasks.  

\section{Model evaluation and selection}


Model selection involves estimating the generalizability of various models to identify the \textit{best} one for a problem \cite{hastie2009elements}.
The estimated generalizability of a model is typically obtained by assessing its predictions on unseen data (\textit{e.g.}, a validation set) from what the model has learned on training data.
The expectation is that a model's performance on unseen data is a reliable indicator of its capacity for generalization.
However, the ability of this performance to be a good indicator of a model's potential depends on the implementation of the evaluation strategy.


An evaluation metric must be specified to describe the quality of a model's predictions.
Various metrics are available to choose from (\textit{e.g.}, accuracy, balanced accuracy, precision, \textit{etc.}), and each paints a distinct picture of how a model performed.
Choosing the right metric requires careful consideration, as each metric carries its own set of assumptions; using an unsuitable metric can lead to the selection of ineffective models \cite{japkowicz2006question,hutchinson2022evaluation,naidu2023review}.
For example, using accuracy over balanced accuracy may result in ineffective models for imbalanced classification problems.
Once an evaluation metric is determined, the evaluation strategy (\textit{e.g.}, cross-validation) for generating a model's performance with the evaluation metric can be defined.
The strategy must be carefully considered, as each balances bias and variance differently \cite{arlot2010surveycv,raschka2018model}.

If the performance scores accurately depict a model's generalizability, selecting the best model for a problem is straightforward.
However, what if multiple models achieve equal performance?
It is well known that overly complex models, such as models with a large number of parameters or a high degree of non-linearity, are prone to overfitting and the curse of dimensionality \cite{bishop2006pattern,hastie2009elements}.
From a practical perspective, simpler models may be required due to hardware limitations, such as those found in mobile phones.
Here, we prioritize simpler models and assume this preference is desired.

\section{TPOT2}


The Tree-based Pipeline Optimization Tool 2 (TPOT2)~\cite{ribeiro2024tpot2} is an AutoML tool that uses genetic programming to evolve machine learning pipelines.
Pipelines are represented as a directed acyclic graph.
Each node in the graph contains one machine learning method and its parameters  (\textit{e.g.}, feature selection and engineering, classification, regression, \textit{etc.}).
A pipeline's graph comprises three node types: leaf, inner, and root.
Data is fed to the leaf nodes, and their output is passed to connecting inner nodes or the root node.
The inner nodes process output from leaf nodes or other inner nodes, and pass their output to connecting inner nodes or the root node.
Pipelines are restricted to a single root node, which can be either a classification or a regression model.


The first generation consists of $pop\_size$ randomly generated pipelines that are evaluated on a set of objective functions.
Invalid pipelines or those exceeding a specified evaluation time limit may emerge, prompting their removal from the population.
Parent selection is used to identify $pop\_size$ parents from the current population.
The parents produce offspring through a combination of mutation and crossover, and offspring are evaluated like their parents.
Survival selection is used to identify $pop\_size$ survivors from both the offspring and the current population. 
The resulting survivors constitute the population for the next generation.
The cycle repeats for a given number of generations, or until sufficient time passes without any progress.


The scores obtained from the objectives functions are used to identify the \textit{best} pipeline among all evaluated valid pipelines.
First, all the best-performing pipelines on the first objective are gathered.
Each following objective is used to filter out pipelines from the current subset that did not perform best on the current objective.
This process continues until all objectives are used.
The remaining pipeline is returned, or one is randomly returned if multiple remain.


\section{Lexicase selection}

Lexicase selection \cite{helmuth2015lexicase} takes a unique approach to identifying parent solutions by using individual performance values from multiple test cases to identify parents.
Generally, lexicase takes a list of solutions and their scores on a set of objectives (test cases) as input. 
To select a parent, the entire list of solutions is placed into a selection pool. 
Next, the set of test cases is randomly shuffled, and each test case is used to remove solutions from the selection pool;
solutions that do not have the best score from the current selection pool on the test case are removed. 
This process repeats until all test cases are used.
If multiple solutions remain, one is randomly selected as a parent.




Lexicase has been modified for TPOT2 with $k$-fold cross-validation in the prior work~\cite{ribeiro2024tpot2}.
Every pipeline has two fitness values: a cross-validated score on the training set and a cross-validated pipeline complexity score. 
Each fitness value is an average of $k$ scores for different folds, where each fold's score is itself an average. 
Lexicase selection, as used in TPOT2, deviates from its initially intended design: consider all particular cases and combinations of cases, and explore their implications independently while not averaging them~\cite{spector2023particularity}. 
Lexidate offers a method for incorporating lexicase selection into evolutionary AutoML that aligns more closely with its original design.

\section{Lexicase-based Validation Strategy}


We combine concepts from model evaluation in machine learning and selection from genetic programming.
In machine learning, a model's generalizability is estimated by the resulting score at the end of evaluation (\textit{e.g.}, a score from $10$-fold cross-validation).
In genetic programming, solutions are evaluated with a fitness function, which assigns a fitness (quality) value used for selection.
Generally, genetic programming focuses on evolving \textit{programs} for a problem where many test cases are needed to measure the quality of a program.
The need for multiple test cases in genetic programming perfectly aligns with the need for a large number of data samples in machine learning.
Lexicase selection's success with these kinds of problems inspired us to use multiple test case performances within TPOT2 to identify parent solutions.



Lexidate splits data into a learning set and a selection set, similar to the holdout method \cite{devroye1979holdout}.
A split is defined by the percentage of data in the learning set and the selection set.
All pipelines in a population are trained on the learning set.
After a pipeline is trained, its predictions on the selection set are graded for correctness.
For example, in a classification task, we assign a value of one for the correct prediction and a value of zero otherwise.
This process creates a vector of selection scores.
Once all pipelines are assigned selection scores, lexicase can use these scores to identify parents.
\section{Methods}


We integrated lexidate within TPOT2 to assess its effect on the evolution of machine learning pipelines across six OpenML classification tasks \cite{feurer2022auto}: 167104, 167184, 167168, 167161, 167185, and 189905.
We describe the experimental setup in the following subsections, but further 
details regarding the configuration of TPOT2 and the six OpenML tasks (including source code) can be found in our supplemental material \cite{supplemental_material}.

\subsection{Evaluation strategy}

We used stratified sampling to partition data for both evaluation strategies at the start of a run.
The resulting data splits remained fixed for an entire run but varied between independent runs.

\subsubsection{$10-$fold cross-validation}
We calculate both an accuracy and a complexity score for each validation fold, then average across each fold's score.
As such, each pipeline will have two scores: cross-validated accuracy and cross-validated complexity.
We count the number of trainable parameters within a pipeline to calculate complexity.
Both scores are used by lexicase to identify parents.

\subsubsection{Lexidate}
We use $90/10$, $70/30$, and $50/50$ splits to generate learning and selection sets.
After a pipeline is trained on the learning set, we calculate its scores on the selection set and record its complexity in terms of the number of trainable parameters.
Note that only graded predictions on the selection set are used by lexicase.

\subsection{Model selection}

The \textit{best} pipeline found throughout the entire run is returned at the end of a TPOT2 run.
This pipeline is then trained on the entire training data and makes predictions on the testing data.
The method to find the best pipeline varies between evaluation strategies.


For 10-fold cross-validation, we collect all pipelines with the highest accuracy.
From this subset, we remove all pipelines that do not have the minimum complexity score found within the remaining pipelines.
If multiple pipelines remain, one is randomly returned. 


Lexidate follows the same procedure, but the filtering objectives are computed differently.
For accuracy scores, we calculate the accuracy of a pipeline's predictions on the selection set.
For complexity scores, we calculate the complexity of a pipeline after training on the learning set.
Both scores are then used to identify the best pipeline. 


\subsection{Experimental design}


Under both evaluation strategies, we evolved populations of $48$ pipelines, with the resulting scores from each strategy used by lexicase to select parents.
We initialized pipelines to have at most ten nodes.
Lexicase identified $48$ parents (with replacement); each parent reproduces asexually and only mutations are applied to offspring.
We ensure no cycles are introduced in a pipeline's graph during mutation.
Moreover, the original pipeline is retained in the following population should a mutation fail for any other reason.
The offspring form the next population, and the cycle repeats for $200$ generations.
We imposed a $30$-minute time limit during pipeline evaluation.
Note that no survival selection is used.


We ran each of the four evaluation strategies on each OpenML task, giving us a total of $24$ treatments.
We performed $40$ replicates for each treatment.
For each replicate, we compared the accuracy and complexity scores of the best pipeline on an unseen test set between the four evaluation strategies.
We used a Kruskal–Wallis test to determine if significant differences occurred between all evaluation strategies on each OpenML task.
For comparisons where differences were detected, we performed a post-hoc pairwise Wilcoxon rank-sum test to identify significant differences between strategy pairings with a Bonferroni correction for multiple comparisons.
We used a significance level of $0.05$ for all tests.
The source code for our experiments, statistics, and visualizations can be found in our supplemental material \cite{supplemental_material}.
All data is available on the Open Science Framework at \href{https://osf.io/mnzjg/}{https://osf.io/mnzjg/}.

\begin{figure}[t!]
\includegraphics[width=\linewidth]{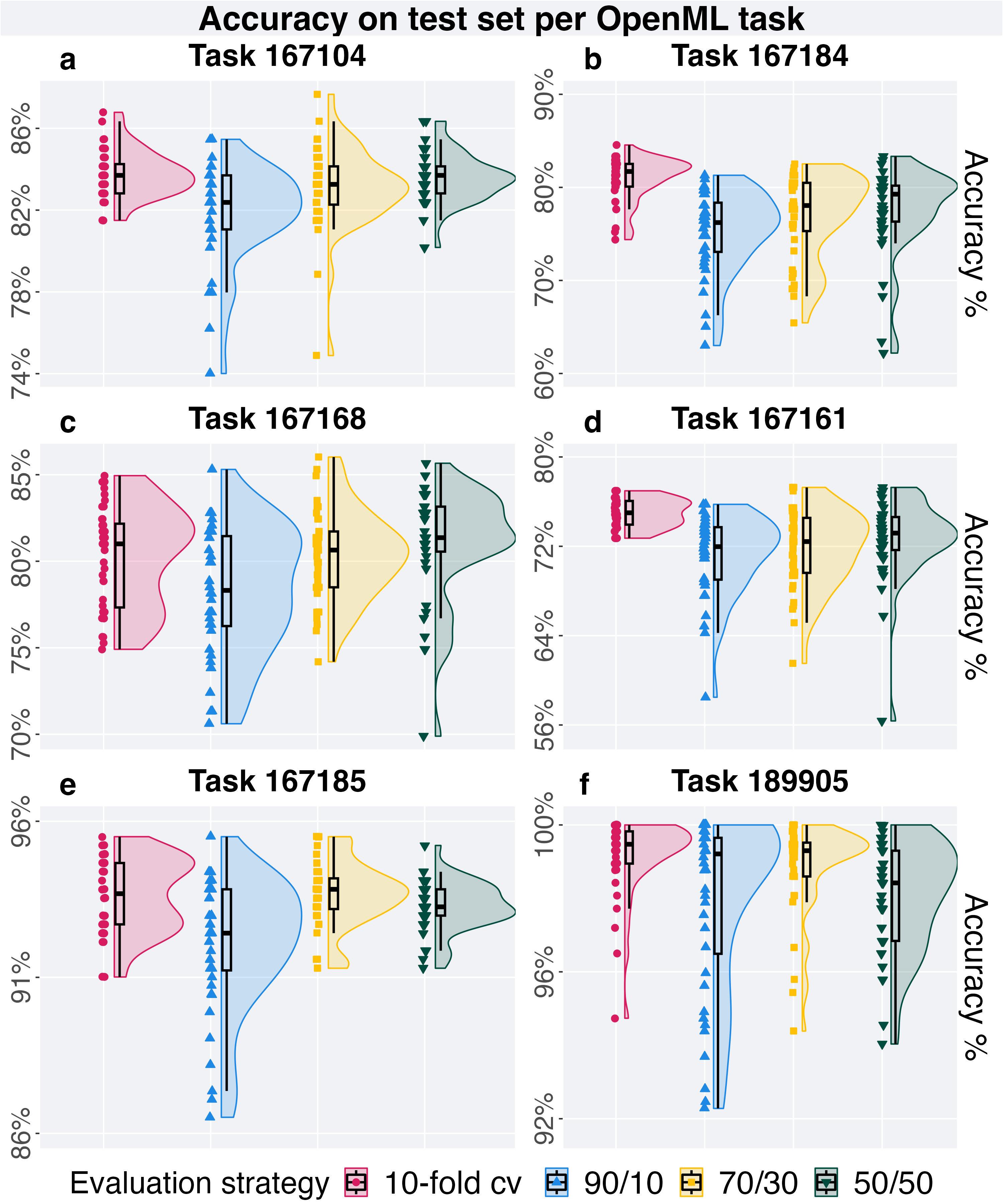}
\caption{Raincloud plots of the accuracy values of the best models returned under different evaluation strategies. Each point in the plot corresponds to a different run.}
\label{fig:res:accuracy} 
\end{figure}


\section{Results}

Figure \ref{fig:res:accuracy} shows the accuracy of the pipeline returned from TPOT2 for each of the four evaluation strategies across the test sets of each OpenML task.
In four of the six tasks, we found no significant differences between 10-fold cross-validation and at least two different lexidate splits (Figure \ref{fig:res:accuracy}a, \ref{fig:res:accuracy}c, \ref{fig:res:accuracy}e, and \ref{fig:res:accuracy}f; Wilcoxon rank-sum test: $p > 0.05$).
For tasks 167184 and 167161, we found significant differences when comparing all lexidate splits to $10$-fold cross-validation (Figure \ref{fig:res:accuracy}b and \ref{fig:res:accuracy}d; Wilcoxon rank-sum test: $p<10^{-3}$).
Table \ref{tab:accuracy:pval} reports all p-values for the accuracy comparisons between the three lexidate splits and the $10$-fold cross-validation. 
Interestingly, each lexidate split evolved pipelines that performed similarly to $10$-fold cross-validation on at least two tasks.
For each lexidate split, the $90/10$ split had no differences on tasks 167168 and 189905, the $70/30$ split had no differences on tasks 167104, 167168, 167185, and 189905, and the $50/50$ split had no differences on tasks 167104, 167168, and 167185.
These results illustrate that lexidate can perform similarly to 10-fold cross-validation, but lexidate's configuration must be considered when solving new problems.





No differences in complexity were detected between all of the evaluation strategies used in this work for tasks 167104, 167168, and 167185 (Kruskal-Wallis test: $p>0.05$).
In the remaining tasks, however, at least one of the lexidate splits evolved pipelines that were less complex than 10-fold cross-validation (Wilcoxon rank-sum test: $p<10^{-2}$).
All lexidate splits evolved less complex pipelines than 10-fold cross-validation on tasks 167184 and 167161 (Wilcoxon rank-sum test: $p<10^{-2}$).
For task 189905, we found that the $90/10$ found less complex pipelines than 10-fold cross-validation (Wilcoxon rank-sum test: $p<10^{-2}$), while the other splits found pipelines with similar complexity (Wilcoxon rank-sum test: $p>0.05$).
These results illustrate that the best pipelines evolved through lexidate are often equally or less complex as those evolved with 10-fold cross-validation.
More details on statistics and all visualizations can be found in our supplemental material \cite{supplemental_material}.

\begin{table}[t]
  \caption{P-values for two-tailed Wilcoxon rank sum test on performances of the best pipeline evolved between three lexidate configurations and 10-fold cross-validation. The last row denotes the number of tasks where no significant differences were detected between a lexidate split and 10-fold cross-validation.}
  \centering
  \begin{tabular}{|c|c|c|c|c|}
    \hline
    \multicolumn{2}{|c|}{} & \multicolumn{3}{c|}{Lexidate split} \\
    \cline{3-5}
    \multicolumn{2}{|c|}{} & 90/10 & 70/30 & 50/50 \\
    \hline
    \multirow{6}{*}{\rotatebox[origin=c]{90}{OpenML task}} & 167104 & $p<10^{-3}$ & $\textbf{p=0.41651}$ & $\textbf{p=1.0}$ \rule{0pt}{2.4ex} \\
    \cline{2-5}
    & 167184 & $p<10^{-8}$ & $p<10^{-4}$ & $p<10^{-3}$ \rule{0pt}{2.4ex} \\
    \cline{2-5}
    & 167168 & $\textbf{p=0.1970}$ & $\textbf{p=1.0}$ & $\textbf{p=1.0}$ \rule{0pt}{2.4ex} \\
    \cline{2-5}
    & 167161 & $p<10^{-7}$ & $p<10^{-4}$ & $p<10^{-3}$ \rule{0pt}{2.4ex} \\
    \cline{2-5}
    & 167185 & $p<10^{-3}$ & $\textbf{p=1.0}$ & $\textbf{p=0.89845}$ \rule{0pt}{2.4ex} \\
    \cline{2-5}
    & 189905 & $\textbf{p=0.25039}$ & $\textbf{p=0.72764}$ & $p<10^{-3}$ \rule{0pt}{2.4ex} \\
    \hline
    \hline
    \multicolumn{2}{|c|}{Summary} & \textbf{2} & \textbf{4} & \textbf{3} \\
    \hline
  \end{tabular}
  \label{tab:accuracy:pval}
\end{table}
\section{Conclusion}

Here, we introduced the lexicase-based validation (lexidate) strategy for model evaluation and selection within TPOT2.
The bottom row in Table \ref{tab:accuracy:pval} counts tasks where no difference in accuracy was detected for the best pipelines between lexidate splits and 10-fold cross-validation.
For lexidate splits, no differences were observed on four tasks with a 70/30 split, three tasks with a 50/50 split, and two tasks with a 90/10 split.
This indicates that the split must be carefully considered for a problem with the amount of available data.
In future work, we plan to investigate if general rules on optimal splits can be established with data set properties (\textit{e.g.}, number of rows and features).
We believe one of the main reasons lexidate performs similarly to 10-fold cross-validation is because it uses individual test cases with no aggregation, a concept called particularity~\cite{spector2023particularity}.

Lexidate is more computationally efficient than 10-fold cross-validation.
A model is trained on $90\%$ of the training data $10$ times with 10-fold cross-validation.
In lexidate, training occurs only once, with the number of samples in the learning set determined by the split configuration.
As such, the lexidate reduces the number of learning calls by a factor of 10, with the time spent learning further reduced by the size of the learning set.

The number of training calls between lexidate and 10-fold cross-validation must also be considered.
As previously mentioned, 10-fold cross-validation uses 10 times more learning calls than lexidate.
Additionally, 10-fold cross-validation makes predictions on all the training samples but lexidate uses predictions only on a subset (\textit{i.e.} the selection set).
Therefore, 10-fold cross-validation provides more snapshots into a pipeline's generalizability compared to lexidate.
One method for keeping comparisons fair between evaluation strategies is to keep the number of training calls consistent between lexidate and 10-fold cross-validation. 
To achieve this, more generations can be given to an evolutionary run with lexidate or more evolutionary runs can be executed.
This idea will be investigated in future work.

\begin{acks}
    We thank the members from the Department of Computational Biomedicine at Cedars-Sinai Medical Center for their helpful comments and suggestions on this work.
    Cedars-Sinai Medical Center provided computational resources through their High Performance Computing clusters.
    The work was supported by NIH grants R01 LM010098 and U01 AG066833
\end{acks}

\bibliographystyle{ACM-Reference-Format}
\bibliography{references,software}

\end{document}